\documentclass[journal]{IEEEtran}
\usepackage{graphicx}

\usepackage{xcolor}
\usepackage[printwatermark]{xwatermark}
\newwatermark*[page=1,color=gray!60,fontfamily=cmr,fontseries=m,scale=0.4,xpos=0,ypos=130]{Preprint in AACD 2021 Workshop Proceedings}

\usepackage{color}

\begin{document}
%
\title{Prospects for Analog Circuits in Deep Networks}
%
%
%

\author{\IEEEauthorblockN{
Shih-Chii Liu$^1$, John Paul Strachan$^2$, and  Arindam Basu$^3$}

\IEEEauthorblockA{$^1$Institute of Neuroinformatics, University of Zurich, Switzerland \\
$^2$ Hewlett Packard Research Labs, Palo Alto, CA, USA \\
$^3$Nanyang Technological University, Singapore and City University of Hong Kong, 
Hong Kong}
\IEEEoverridecommandlockouts
}

\maketitle

\thispagestyle{empty}

\begin{abstract}
\boldmath
Operations typically used in machine learning algorithms (e.g. adds and soft max) can be implemented by compact analog circuits. Analog Application-Specific Integrated Circuit (ASIC) designs that implement these algorithms using techniques such as charge sharing circuits and subthreshold transistors, achieve very high power efficiencies. 
With the recent advances in deep learning algorithms, focus has shifted to hardware digital accelerator designs that implement the prevalent matrix-vector multiplication operations. Power in these designs is usually dominated by the memory access power of off-chip DRAM needed for storing the network weights and activations. 
Emerging dense non-volatile memory technologies can help to provide on-chip memory and analog circuits can be well suited to implement the needed multiplication-vector operations coupled with in-computing memory approaches.
 This paper presents a brief review of analog designs that implement various machine learning algorithms. It then presents an outlook for the use of analog circuits in low-power deep network accelerators suitable for edge or tiny machine learning applications.

\end{abstract}


%

%

\section{Introduction}
Machine-learning systems produce state-of-art results for many applications including data mining and machine vision. They extract features from the incoming data on which decisions are made, for example, in a visual classification task. Current deep neural network approaches in machine learning (ML)~\cite{lecun2015deep} produce state-of-art results in many application domains including visual processing (object detection~\cite{review_alexnet}, face recognition~\cite{review_face} etc), audio processing (speech recognition~\cite{review_speech1}, keyword spotting etc), and natural language processing~\cite{review_nlp}, to name a few. This improvement in algorithms and software stack has eventually led to a drive for better hardware that can run such ML workloads efficiently. 

In recent times, edge computing has become a big research topic and edge devices that first process the local input sensor data are being developed within different sensor domains.  Many current deep network hardware accelerators designed for edge devices are implemented through digital circuits. Less explored are analog/mixed-signal designs that can provide high energy-efficient implementations of deep network architectures. 
Custom analog circuits can exploit the physics of the transistors in implementing computational primitives needed in deep network architectures. For example,  operations such as summing can be implemented by simpler  analog transistor circuits rather than digital circuits. These designs can provide better area and energy efficiencies than their digital counterparts in particular for edge inference applications which require only small networks. With the increasing availability of dense new non-volatile memory (NVM) technology, the computation can be co-localized with memory, therefore further savings in power can be obtained by the reduction of off-chip memory access. 

This article reviews analog and mixed-signal hardware chips that implement ML algorithms including the  deep neural network architectures. We discuss the advantages of analog circuits and also the challenges of using such circuits. While other reviews have focused on analog circuits for on-chip learning~\cite{jetcas_review}, we focus more on implementations of generic building blocks that can be used in both training and inference phases.

The paper organization is as follows: First, Sec.~\ref{sec:a-networks} provides a brief  review on analog circuits used for neuromorphic computing and how these circuits can implement computational primitives useful for neural processor designs and ML algorithms. It is followed by Sec.~\ref{sec:vmm} that describes basic operations (such as vector-matrix multiplications) being accelerated by analog building blocks, and Sec. \ref{sec:crossbars} that reviews the field of non-volatile memory technologies for deep network accelerators.
Section \ref{sec:analog-dnn} reviews recent Application-Specific Integrated Circuits (ASICs) that implement certain ML algorithms. Finally, we end the paper with a discussion about future prospects of analog-based ML circuits (e.g. in-memory computing).
 

\section{Review of circuits for analog computing}
\label{sec:a-networks}
With great advances made in digital circuit design through the availability of both software and hardware design tools, analog circuits have taken a back seat in mainstream circuits and are now primarily used in high speed analog domains such as RF, power regulation, PLLs, and interface of sensors and actuators (e.g. in the analog-digital converter (ADC) readout circuits of analog microphone or biochemical sensors). 

When it was proposed that the exponential properties of running the transistors in subthreshold can be useful for emulating the structure of nervous system in the field of neuromorphic engineering~\cite{carver1989analog}, analog circuits became popular for emulating biological structures like the retina~\cite{mead1988silicon}, implementing generalized smoothing networks~\cite{liu1989generalized,harrisRes91}, and both simplified and complex biophysical models of biological neurons. The exponential properties of a subthreshold transistor also make it easier to design analog VLSI circuits for implementing basic functions used in many mixed-signal neuromorphic systems such as the sigmoid, similarity~\cite{delbrueck1993bump}, charge-based vector-matrix calculations~\cite{genovVM2001}, and highly distributed operations such as the winner-take-all~\cite{liu2002analog}.


Analog ML designs using these basic functions included a low-power analog radial basis function programmable system~\cite{peng2007analog};  
a sub-microWatt Support Vector Machine classifier design~\cite{ChakrabarttySVM2007} 
with a computational efficiency of $\sim$ 1 TOp/s/W; and the analog deep learning feature extractor system~\cite{Holleman2015} that processes over 8K input vectors per second and achieves 1 TOp/s/W. The power efficiency of ~\cite{Holleman2015} is achieved by running the transistors in subthreshold and using floating-gate non-volatile technology for both storage, compensation and reconfiguration.

Even though analog neural processor circuits were already reported in the 1990s~\cite{masa1994high}, they had taken a back seat partially because of the ease of doing digital designs; and the care needed for good matching in analog circuits. Although mismatch is usually reduced by using larger transistors, there are circuit strategies to minimize the mismatch effect, for example, by techniques to simplify the circuit complexity (Sec.~\ref{sec:vmm}), floating-gate techniques to compensate for the mismatch, and training methods for deep networks to account for this mismatch as will be discussed in Sec.~\ref{sec:analog-dnn}.

\begin{figure}[!ht]
\centering
\includegraphics[width=0.5\textwidth]{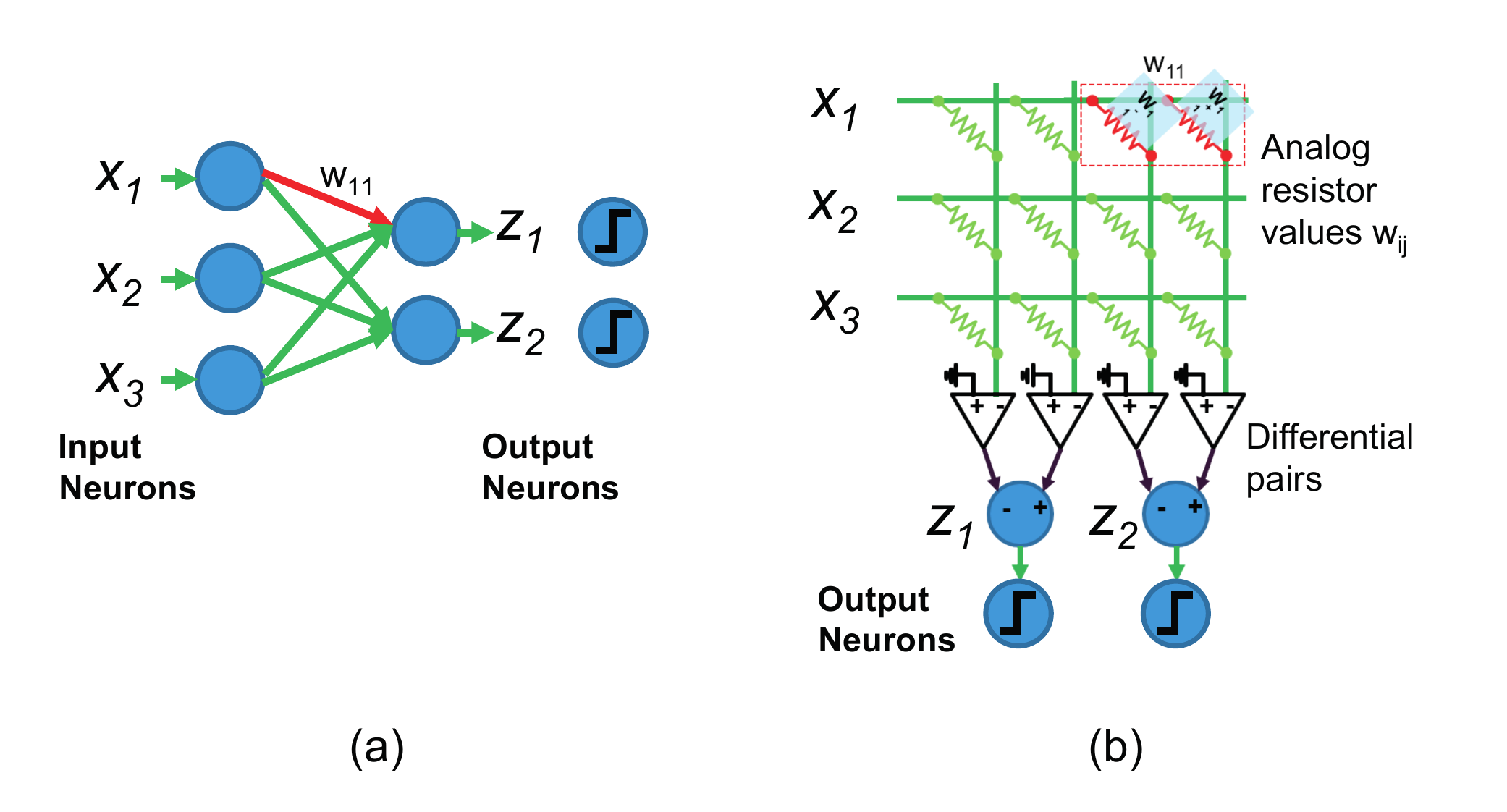}
\caption{Simple two-layer Artificial Neural Network (ANN) using a non-linear transfer function (a) and an implementation using a resistive crossbar array with differential weights: two non-negative resistors are used to represent a bipolar weight (b).}
\label{fig:crossbar}
\end{figure}

\begin{figure}[!ht]
\centering
\includegraphics[width=0.35\textwidth]{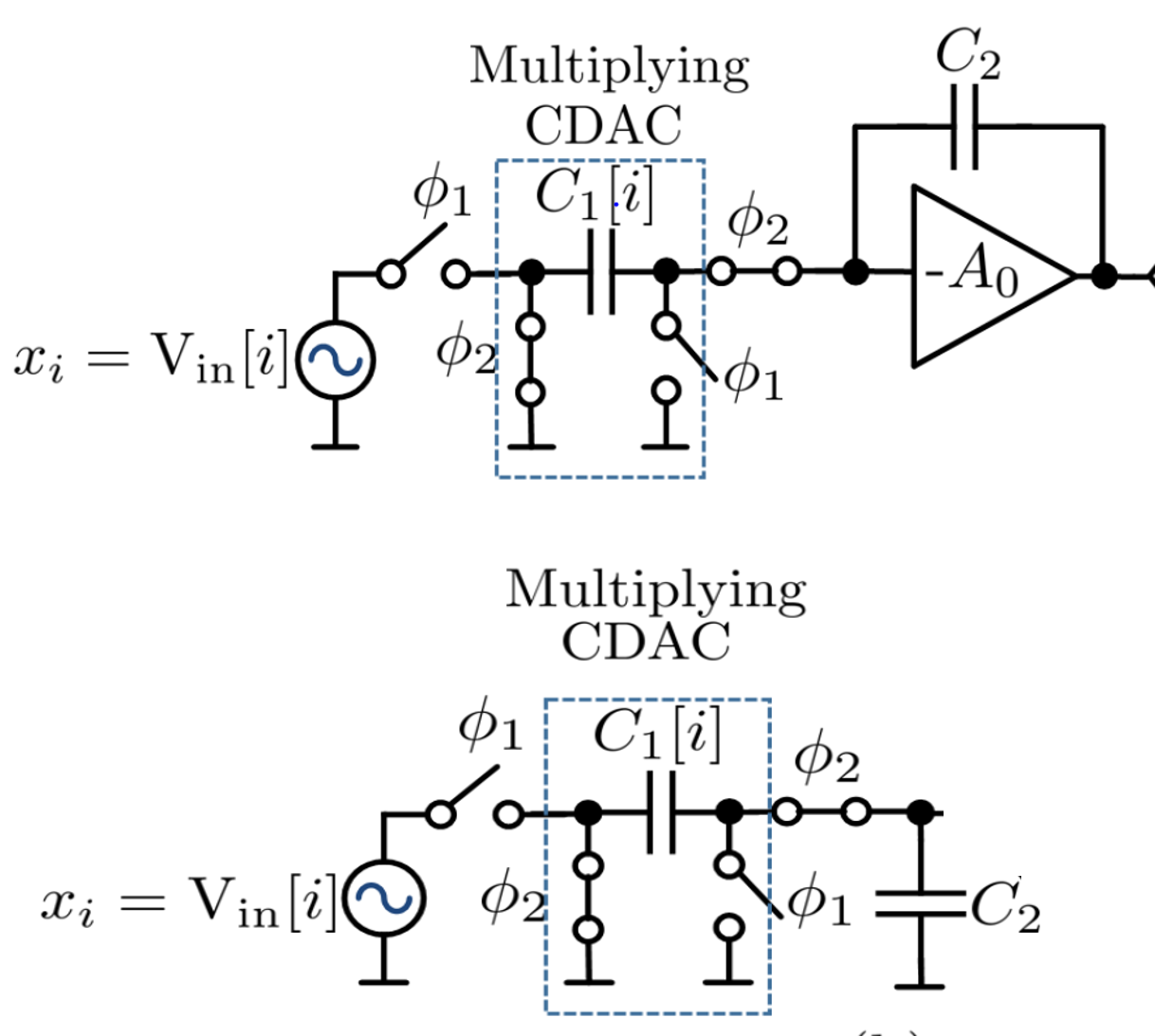}\\
(a)\\
\includegraphics[width=0.45\textwidth]{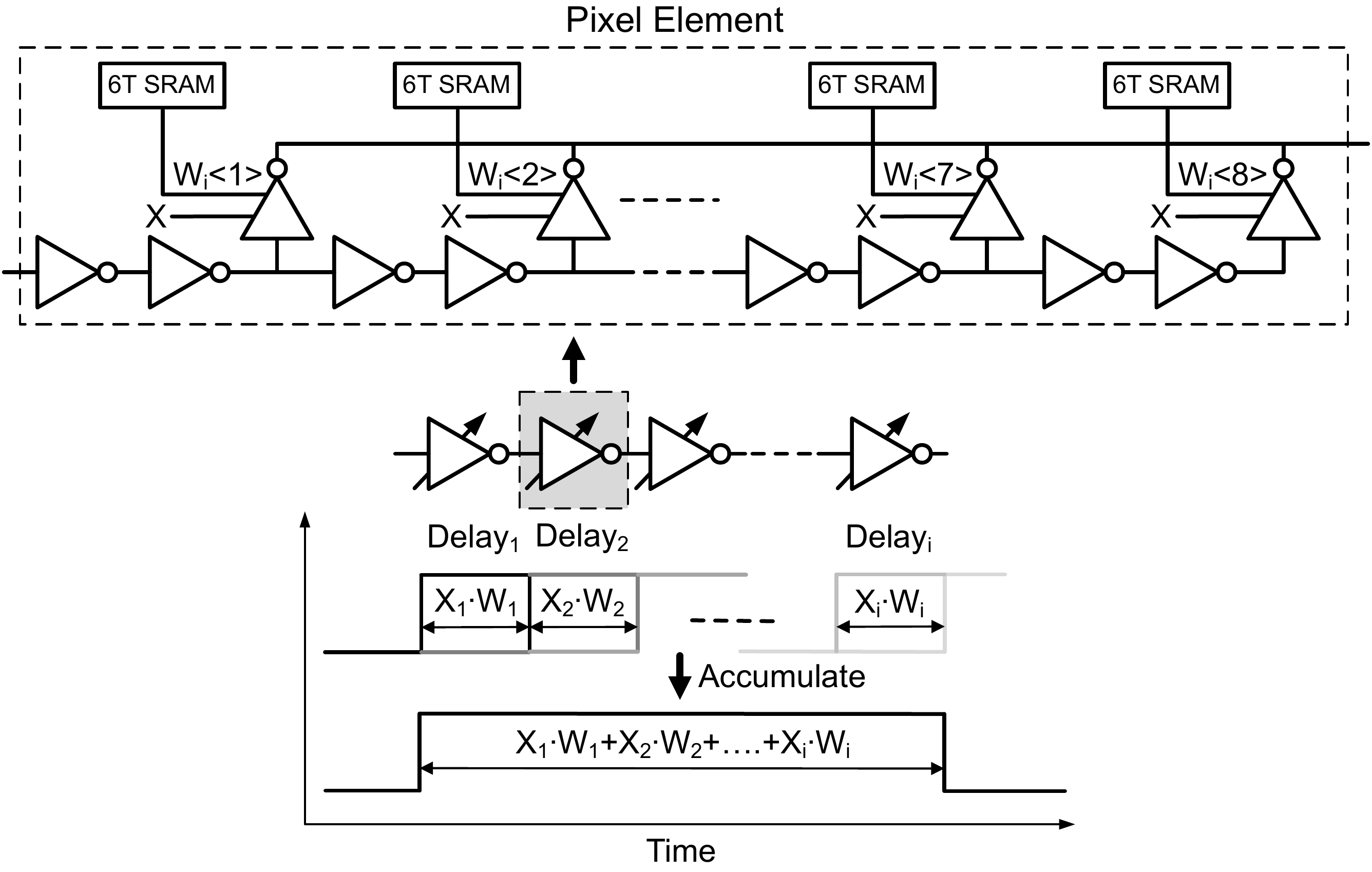}\\
(b)\\
\includegraphics[width=0.4\textwidth]{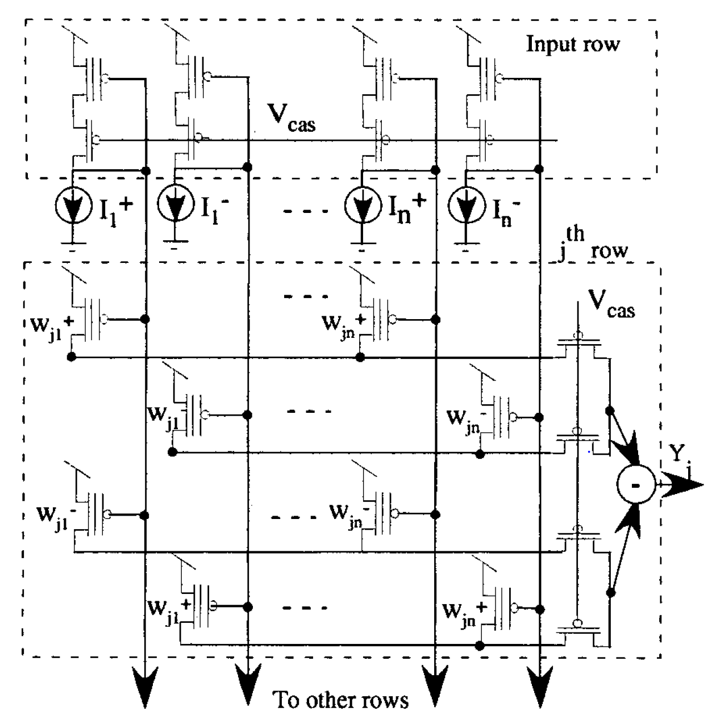}\\
(c)
\caption{Three modes of matrix-vector multipliers using analog circuits: (a) Charge (b) Time and (c) Current. Figures adapted from \cite{charge_jssc}, \cite{time_asscc} and \cite{current_cicc}.
}
\label{fig:analog-vmm-styles}
\end{figure}

\section{Analog circuits for matrix-vector multiplication}
\label{sec:vmm}

In  a multi-layered neural network, the output of a neuron $i$ in layer $l$ is given by $y_i^l = g(z_i^{l}) = g\smash{(\sum_j w_{ij} x_j^l)}$, where $g()$ is a nonlinear function and the $j$-th input to the $l$-th layer $x_j^l$ is the $j$-th output from the previous layer (Fig.~\ref{fig:crossbar} (a)).
The basic operations: summation, multiplication by scalars, and simple nonlinear transformations such as sigmoids can be implemented in analog VLSI circuits very efficiently. Dropping the index $l$ denoting layer, we can write the most computationally intensive part as a matrix-vector multiplication (MVM) as follows:
\begin{equation}
    \overline{z} = \mathbf{W}\overline{x}
\end{equation}
where $\overline{x}$ is the vector of inputs to a layer, \textbf{W} denote the matrix of weights for the layer and $\overline{z}$ denotes the vector of linear outputs which is then passed through a nonlinear function $g()$ to produce the final neuronal output.  Figure \ref{fig:analog-vmm-styles} illustrates how the key core computation of weighted summation can be performed in three modes: (a) Charge, (b) time and (c) current. Further  storage of weight coefficients can be performed in a volatile or non-volatile manner. More details about these different possibilities are described next.
 \paragraph{Charge}
 Figure \ref{fig:analog-vmm-styles}(a) depicts a capacitive digital analog converter (CDAC) based charge mode MVM circuit. In this circuit, the weight is stored in the form of binary weighted capacitor sizes (shown as $C_1[i]$ in the figure) and hence weighting of input happens naturally by charging the desired capacitor value with the input voltage. Summation happens through the process of charge redistribution among capacitors via switched capacitor circuit principles. The charge redistribution can be more accurate if done by using an amplifier in a negative feedback configuration (as shown on top of Fig. \ref{fig:analog-vmm-styles}(a))--this is referred to as an active configuration~\cite{charge_jssc,charge_isscc}. On the other hand, the passive configuration without an amplifier (shown below the active one) has the benefit of high speed operation due to the absence of settling time limitations from the amplifier. However, it was shown in \cite{charge_jssc} that the errors in the passive multiplication amount to a change in the coefficients of the weighting matrix $\textbf{W}$ and can be accounted for. A disadvantage of this approach is that it requires $N$ clock cycles to perform a dot product on two $N$ dimensional vectors. Static random access memory (SRAM) is used to store the weights--hence, it suffers from volatile storage issues. A 6-b input, 3-b weight implementation of this approach achieved $\approx 7.7 -8.7$ TOp/s/W energy efficiency at clock frequencies of $1-2.5$ GHz clock frequency~\cite{charge_jssc}. Other recent implementations using this approach are reported in \cite{cdac_andreas}  and \cite{gert_cicc}.
 
 \paragraph{Time}
Another  way of implementing addition in the analog domain is by using time delays. Figure \ref{fig:analog-vmm-styles}(b) shows how addition can be implemented through the accumulation of delays in cascaded digital buffer stages~\cite{time_asscc}. By modifying the delay of each stage according to the weight, a weighted summation can be achieved. Figure \ref{fig:analog-vmm-styles}(b) shows one example of a delay line based implementation of this approach. Here, the weight storage is in volatile SRAM cells. The authors used a thermometer encoded delay line as an unit delay cell to avoid nonlinearity at the expense of area. A 1-b input, 3-bit weight architecture achieved very high energy efficiency of $\approx 105$ TOp/s/W at $0.7$ V power supply~\cite{time_asscc}.
 
 \paragraph{Current}
 By far, the most popular approach for analog VMM implementation is the current mode approach. This architecture offers flexibility in choosing the way input is applied (encoding magnitude in current~\cite{chen2019elm}, voltage~\cite{rram_lstm} or as pulse-width of a fixed amplitude pulse~\cite{marinella2018multiscale}) and non-volatile weight storage (Flash~\cite{vmm_strukov}, PCM~\cite{suri2011phase}, RRAM~\cite{rram_lstm}, MRAM~\cite{grollier2016spintronic}, Ferroelectric FETs~\cite{jerry2017ferroelectric}, ionic floating-gate~\cite{fuller2019parallel}, transistor mismatch~\cite{yaobasu2016}, etc). An example implementation using standard CMOS compatible Flash transistors shown in Fig. \ref{fig:analog-vmm-styles}(c) was presented as early as 2004~\cite{current_cicc}. 
 
 In these approaches, a crossbar array of NVM devices are used to store the weights $\textbf{W}$. In the case of Flash, the inputs $x_j$ can be encoded in current magnitude ($I_j^+ - I_j^-$) and presented along the columns (word-lines) while current summation occurs along the rows (bit-lines) by virtue of Kirchhoff's Current Law (KCL) to produce the output $z_j$. Here, the weighting happens through a sub-threshold current mirror operation of the input diode-connected Flash transistor and the NVM device in the crossbar with weight being encoded as charge difference between the two devices. Since then, several variants of these Flash-based VMM architectures have been published using amplifier based active current mirrors~\cite{vmm_jssc_my} for higher speed, source coupled VMM for higher accuracy~\cite{vmm_Craig} and special Flash process based VMM~\cite{vmm_strukov} with promise of scaling down to 28nm. These approaches generally achieve energy efficiencies in the range of $5-10$ TOp/s/W.
 
 An interesting alternative is to use the threshold voltage mismatch in-built in transistors as a weight quantity. This leads to ultra-compact arrays for VMM operation~\cite{chen2019elm}. However, since these weights are random, they may only be used to implement a  class of randomized neural networks such as reservoir computing, neural engineering framework, extreme learning machines etc. While these networks are typically only two layers deep, they have advantages of good generalization and quick retraining. Such approaches have been used for brain-machine interfaces~\cite{chen2016elm}, tactile sensing~\cite{tactile_elm} and image classification~\cite{patil_neurocomputing}.
 
 The recent trend in this architecture is to use resistive memory elements, sometimes dubbed memristors, as the NVM device in the crossbar. They are less mature but offer advantages of longer retention, higher endurance, low write energy, and promise of scaling to sizes smaller than Flash transistors. They have also been shown to support back-propagation based weight updates with small modifications to the architecture. Due to their increasing popularity in deep network implementations, we will discuss them further in the next section.

\section{Non-volatile resistive crossbars}
\label{sec:crossbars}

\begin{figure}[!ht]
\centering
\includegraphics[width=0.4\textwidth]{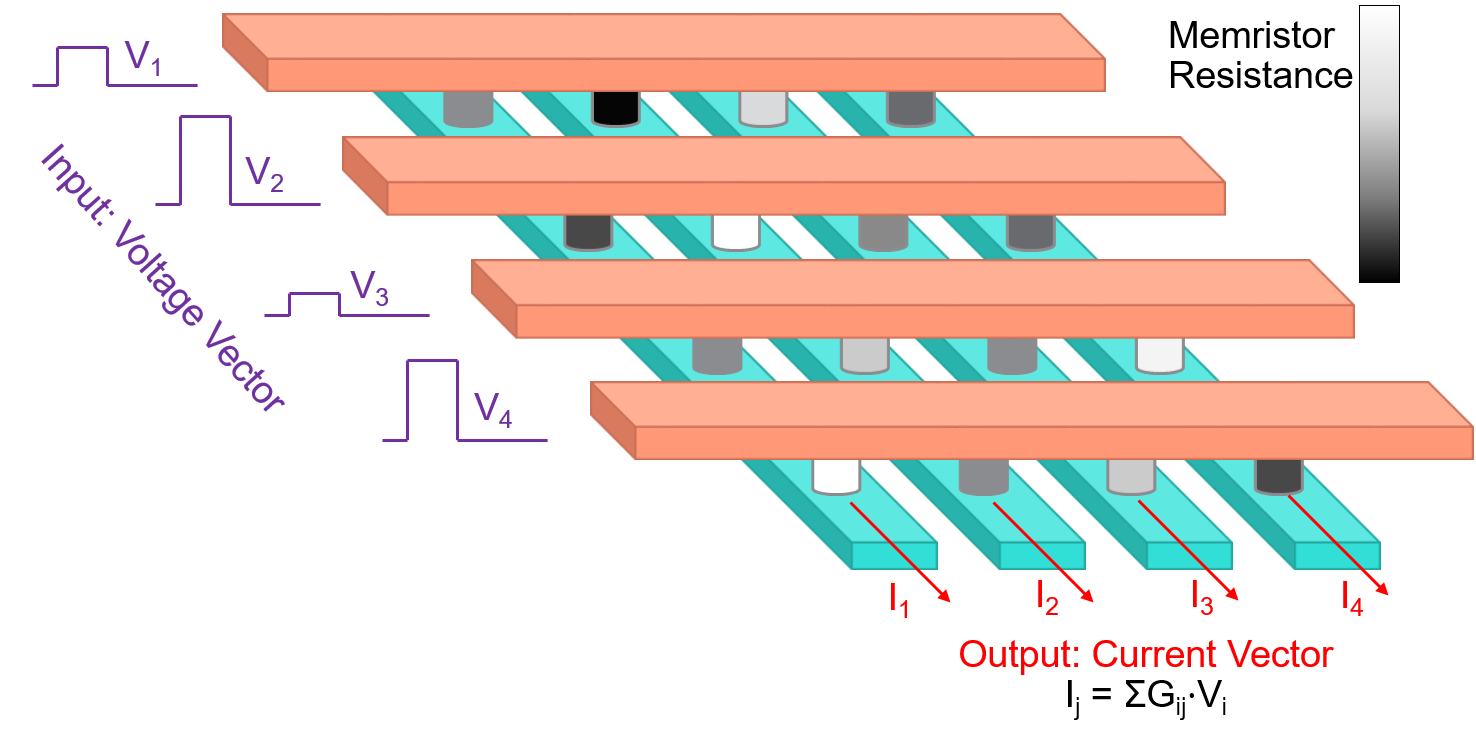}
\caption{Performing vector-matrix multiplication in non-volatile resistive (memristive) crossbar arrays. The input can be encoded in voltage amplitude with fixed pulse width, as shown here, or other approaches such as fixed amplitude and variable pulse width. At each crossbar junction, a variable resistor is present representing a matrix weight element, modulating the amount of current injected into the column wires through Ohm's law. On each column, the currents from each row are summed through Kirchhoff's current law, yielding the desired output vector representing the vector-matrix product. Pairs of differential weights can be utilized as in Fig.~\ref{fig:crossbar}(b) to allow bipolar representations. }
\label{fig:memristor}
\end{figure}

A popular approach for performing the matrix-vector multiplications in ML - and especially in the deep neural networks of today - is to leverage emerging non-volatile memory technologies such as phase change (PCM),  oxide-based resistive RAM (RRAM or memristors), and spin-torque magnetic technologies (STT-MRAM). The equivalent of an ANN implemented in a resistive memory array is shown in Fig.~\ref{fig:crossbar} (b). As further illustrated in Fig.~\ref{fig:memristor}, a dense crossbar circuit is constructed with these resistive elements or memristors at every junction. Each resistor is analog tunable to encode multiple bits of information (from binary to over seven bits~\cite{alibart2012high, hu2018memristor}). Most importantly, the devices retain this analog resistive programming in a non-volatile manner. This is appropriate for encoding the heavily reused weights in deep neural networks, such as convolution kernels or fully connected layers. And, as noted earlier, this allows for the dominant vector-matrix computations to be performed in-memory without the costly fetching of neural network (synaptic) weights.

To take full advantage of analog non-volatile crossbars in deep neural networks, it is necessary to devise a larger architecture that supplements the vector-matrix computations performed in-memory, with a range of additional data orchestration and processing performed by traditional digital circuits. These include operations such as data routing, sigmoidal or other nonlinear activation functions, max-pooling, and normalization. Several prior works have explored this design-space~\cite{shafiee2016isaac, song2017pipelayer} for resistive crossbars, with the inclusion of schemes to encode arbitrarily high precision weights across multiple resistive elements ("bit-slicing") and the overhead in constantly converting all intermediate calculations performed within analog resistive crossbars back into the digital domain to avoid error accumulation in deep networks of potentially tens of layers. An architecture and performance analysis was first done for CNNs~\cite{shafiee2016isaac}, and then extended to nearly all modern deep neural networks including flexible compiler support~\cite{ankit2019puma}.

To date, non-volatile resistive crossbars have been explored for both the inference and training modes of deep neural networks. In the case of inference, the main advantage from NVM comes from reduced data-fetching and movement, while reprogramming of the crossbars is infrequent, requiring lower endurance from the resistive technology (PCM or RRAM), but potentially higher retention and programming yield accuracy~\cite{yang2013memristive}. To handle device faults, programming errors, or noise, novel techniques for performing error-detection and correction for the computations performed within these crossbar arrays have been devised recently~\cite{roth2018fault, roth2020analog2} and experimentally demonstrated~\cite{li2020analog}.  On the other hand, implementation of training in NVM crossbars~\cite{burr2015experimental, prezioso2015training,SebastianCIM2019} can take advantage of efficient update schemes for tuning each of the resistive weights in parallel (e.g., "outer product" updates such as in~\cite{agarwal2016energy}). In turn, the implementation of neural network training puts a larger burden on the underlying NVM technology from an endurance perspective, and requirements for a symmetric and linear change in the device resistance when programmed up or down. While the technology remains highly promising, the current state of resistive NVM suffers from large programming asymmetries and uniformity challenges. As the technology continues to mature, some current approaches are able to mitigate these setbacks by complementing the resistive technology with short term storage in capacitors~\cite{ambrogio2018equivalent}, or operating in a more binary mode~\cite{chen2019cmos}, but with cost of either area or prediction accuracy. Nonetheless, practical demonstrations using resistive crossbar networks have already included, to name just a few, image filtering and signal processing~\cite{li2018analogue}, fully connected and convolutional networks~\cite{prezioso2015training,hu2018memristor, ambrogio2018equivalent}, LSTM recurrent neural networks~\cite{rram_lstm}, and reinforcement learning ~\cite{wang2019reinforcement}.

\begin{figure}[ht!]
\centering
\includegraphics[width=0.45\textwidth]{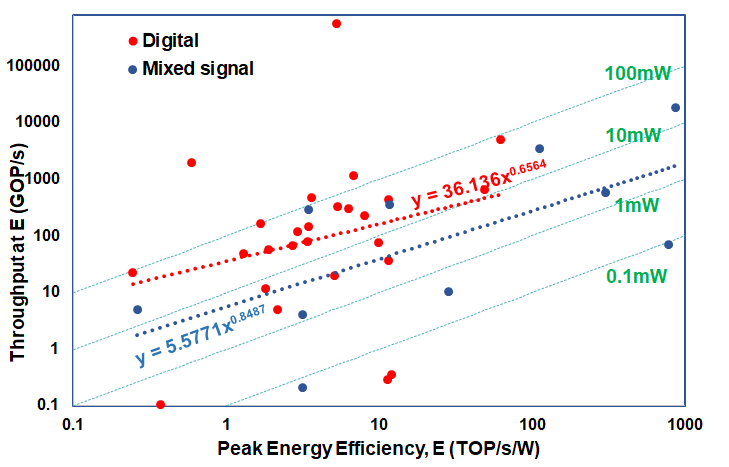}\\
(a)\\
\includegraphics[width=0.45\textwidth]{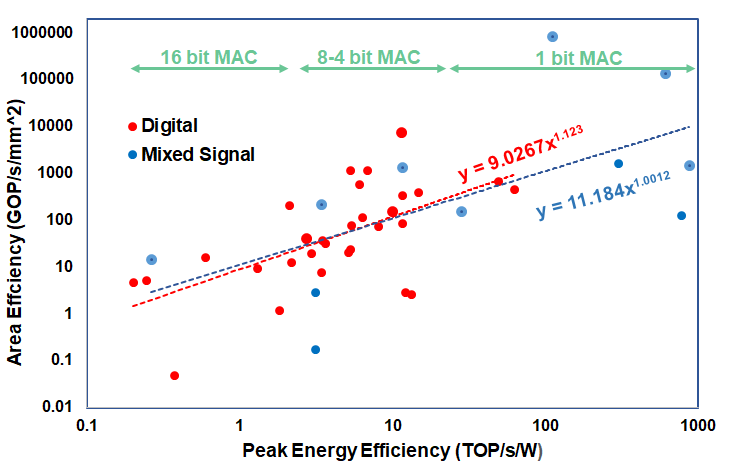}\\
(b)
\caption{
Machine learning hardware trends: Peak energy efficiency vs throughput (a)  and area efficiency (b) for recent ASIC implementations reported in ISSCC, SOVC, and JSSC (see main text).
}
\label{fig:asic_trends}
\end{figure}

\section{Future of analog deep neural network architectures}
\label{sec:analog-dnn}

Analog circuits can be more compact than digital circuits, especially for certain low-precision operations like the add operation needed in neural network architectures or the in-memory computing circuits for summing up weighted currents~\cite{chandrakasan2018conv}. While mismatch variations could be a feature for certain network architectures like the Extreme Learning Networks, they need to be addressed in analog implementations of deep networks by using design techniques such as the schemes described in Sec.~\ref{sec:vmm} or through training methods that account for the statistics of the fabricated devices.

The work of ~\cite{binas2018analog} shows that it is possible to use neural network training methods as an effective optimization framework to automatically compensate for the device mismatch effects of analog VLSI circuits. The response characteristics of the individual VLSI neurons are added as constraints in an off-line training process, thereby  compensating for the inherent variability of chip fabrication and also taking into account the particular analog neuron's transfer function achievable in a technology process. 
The measured inference results from the fabricated analog ASIC neural network chip~\cite{binas2018analog} matches that from the simulations, while offering lower power consumption over a digital counterpart. A similar retraining scheme was used to correct for memristor defects in a memristor-based synaptic network  trained on a classification task~\cite{liu2017rescuing}.

\subsection{Trends in machine learning ASICs}

To understand the energy and area efficiency trends in recent ASIC designs of ML systems, we present a survey~\cite{asilomar2019} of papers published in IEEE ISSCC and IEEE SOVC conferences from 2017-2020 (similar to the ADC survey in \cite{Survey_adc}), and 
in the IEEE Journal of Solid-state Circuits (JSSC).

Using the data from \cite{Survey_excel}, we show two plots. The first plot shows the throughput  \textit{vs.} peak energy efficiency tradeoff for digital and mixed-signal designs (Fig.~\ref{fig:asic_trends}(a)). It can be seen that while throughput is comparable for analog and digital approaches, the energy efficiency for mixed-signal designs is superior to their digital counterparts. Figure \ref{fig:asic_trends}(b) plots the area efficiency vs peak energy efficiency for these designs and further classifies the digital designs according to the bit-width of the multiply-accumulate (MAC) unit. It can again be seen that the mixed-signal designs have better tradeoff than the digital counterparts. Furthermore, as expected, lower bit widths lead to higher efficiencies in terms of both area and energy. Lastly, the efficiency of mixed-signal designs seem significantly better than digital designs only for these with very low-precision 1-bit MACs. 

\section{Conclusion}
\label{sec:conclusion}

While  deep network accelerator designs are implemented primarily using digital circuits, in-memory computing systems can benefit from analog and mixed-signal circuits for niche ultra low-power edge or biomedical applications. The peak energy efficiency  trends in Fig.~\ref{fig:asic_trends} show that  mixed-signal designs can be competitive at lower-bit precision parameters for network accelerators. 
Some of the analog machine learning circuit blocks such as the sigmoid and the winner-take-all designs can implement directly the sigmoid transfer function of the neuron, and the  soft-max function of the final classification layers of a deep network. It will be interesting to see how these circuits will be incorporated further into in-memory computing systems in the future. 


\section{Acknowledgment}

The authors would like to thank Mr. Sumon K. Bose and Dr. Joydeep Basu for help with figures; and T. \,Delbruck for comments. 

\bibliographystyle{IEEEtran}

\bibliography{bibliography}

\ifCLASSOPTIONcaptionsoff
  \newpage
\fi

\end{document}